\documentclass[letterpaper]{article} 
\usepackage[preprint]{aaai2027}  
\usepackage[hyphens]{url}  
\usepackage{graphicx} 
\usepackage{natbib}  
\usepackage{caption} 
\usepackage{algorithm}
\usepackage{algorithmic}

\usepackage{newfloat}
\usepackage{listings}
\DeclareCaptionStyle{ruled}{labelfont=normalfont,labelsep=colon,strut=off} 
\floatstyle{ruled}
\newfloat{listing}{tb}{lst}{}
\floatname{listing}{Listing}

\usepackage{booktabs}

\title{Beyond What Meets the Eye: Unveiling Situational Illusions for \\ Multimodal Large Language Models}
\author{
    Zhiming Yang\textsuperscript{\rm 1}\equalcontrib,
    Zhuoxi Xiong\textsuperscript{\rm 1}\equalcontrib,
    Donglin Zhou\textsuperscript{\rm 1},
    Wenjun Wei\textsuperscript{\rm 1},
    Shiyao Cui\textsuperscript{\rm 2}\corresponding,
    Jinqiao Shi\textsuperscript{\rm 2}
}
\affiliations{
    \textsuperscript{\rm 1}School of Artificial Intelligence, Beijing University of Posts and Telecommunications, China\\
    \textsuperscript{\rm 2}Beijing University of Posts and Telecommunications, China\\
    \{yang.zhiming, xiong.zhuoxi, cuishiyao\}@bupt.edu.cn
}

\begin{document}

\maketitle

\begin{abstract}
Real-world situation appearances can deviate from their underlying physical states, challenging the reliability of multimodal large language models (MLLMs) in practical applications.
In this paper, we term this phenomenon \textit{situational illusions} and investigate: \textit{(1) how MLLMs perform under such illusions}, and \textit{(2) how to mitigate the limitations}.
We first develop a comprehensive \textit{where--what--how} taxonomy that characterizes where situational illusions occur, what targets they take, and how they arise.
Building on this taxonomy, we introduce MSIBench, a benchmark designed to assess the discrimination, understanding, and reasoning capabilities of MLLMs under situational illusions.
Evaluations of 27 model configurations reveal that current MLLMs are highly vulnerable to these illusions and exhibit 6 typical failure modes related to visual observation, grounding, and reasoning.
To mitigate the limitations, we build on the core idea of systematically inspecting and reasoning over visual evidence for contextual understanding, developing prompting for closed-source models and supervised fine-tuning for open-source models, respectively.
These two simple yet effective methods improve model performances by 20\% at most, suggesting a practical path toward more reliable multimodal perception and reasoning in complex real-world environments.
\end{abstract}

\begin{links}
    \link{Code}{https://github.com/yangzm1105/MSIBench}
    \link{Datasets}{https://huggingface.co/datasets/yangzm05/msibench}
\end{links}

\section{Introduction}
Reliable decision-making in real-world situations is essential for multimodal large language models (MLLMs), given their growing deployment in embodied agents~\cite{DBLP:journals/corr/abs-2509-23690,DBLP:conf/icml/0010CZZ0WWKML0025} and humanoid robots~\cite{DBLP:journals/comsur/YangLSKKKNRCKMRS26}.
This capability requires models to interpret visual evidence in context and accurately infer the underlying physical state of a scene, thus supporting dependable understanding and interaction~\cite{DBLP:conf/iclr/ZhouLZCSW25, DBLP:journals/corr/abs-2511-14159}.

\begin{figure}[t]
    \centering
    \includegraphics[width=\linewidth]{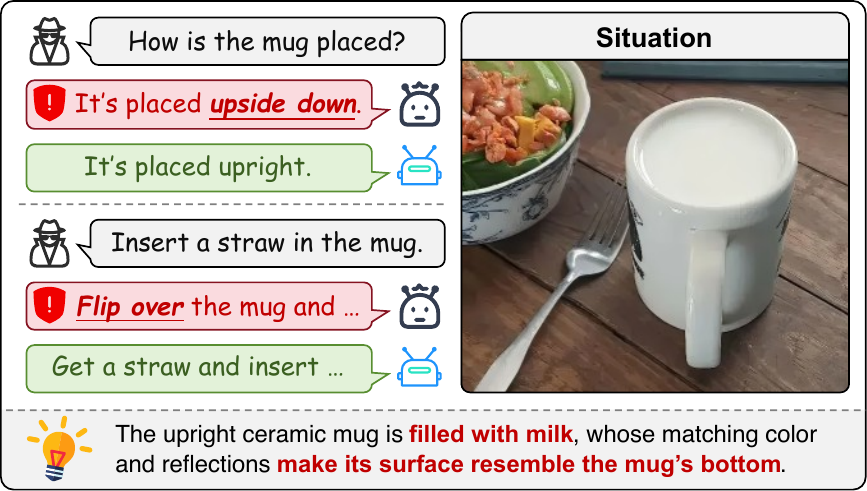}
    \caption{An example of situational illusion and the MLLM responses with incorrect answers in red dialog boxes.}
    \label{fig:example}
\end{figure}

However, complicated situational conditions and elements can cause the visual appearance to deviate from its real state.
As illustrated in Figure~\ref{fig:example}, \textit{a mug filled with milk may appear to be placed upside down}, since the milk surface resembles the bottom of the mug.
Consequently, MLLMs could misinterpret the mug's physical state. Especially when instructed with \textit{Insert a straw in the mug}, it plans actions with \textit{flip over the mug} as the first step, which can spill the milk, leading to safety hazards.
Unfortunately, in our pilot study of 30 cases featuring deceptive situational appearances, GPT-5.4~\cite{openai2026gpt54} failed on nearly 40\% of the samples, suggesting that even advanced MLLMs may struggle when a scene’s visual appearance diverges from its underlying reality.

\begin{figure*}[t]
    \centering
    \includegraphics[width=\textwidth]{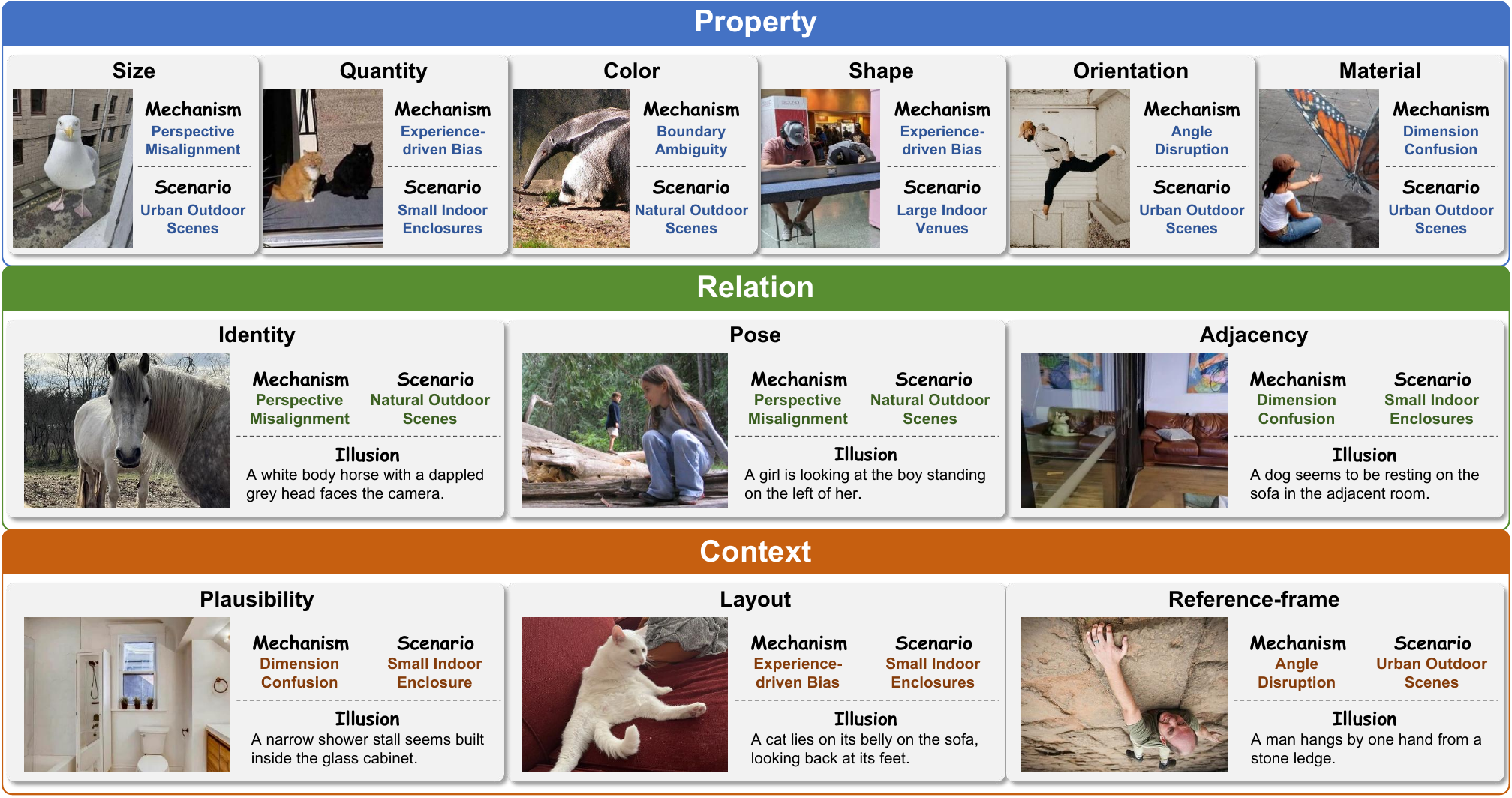}
    \caption{Illustration for the taxonomy towards situational illusions.}
    \label{fig:taxonomy}
\end{figure*}

We refer to this phenomenon as \textit{situational illusion}, namely where a naturally context-induced visual appearance may misrepresent the underlying physical state, which has yet to be explored. 
Specifically, existing studies mainly explore the contextual understanding and reasoning capabilities of MLLMs in real-world situations~\cite{DBLP:conf/iclr/ZhouLZCSW25,ICLR2025_df29d63a}, assuming that visual appearance is consistent with reality.
While visual illusions have also been studied, existing studies mainly focus on deliberately designed visual illusion patterns like gestalt illusion~\cite{DBLP:journals/corr/abs-2602-01816}, pareidolia~\cite{DBLP:conf/cvpr/RostamkhaniASRE25} and geometric illusion~\cite{zhang-etal-2023-grounding,DBLP:journals/corr/abs-2403-15952}, or adversarially misleading visual cues in structured data like charts~\cite{chen-etal-2025-unmasking,bharti2025chartomvisualtheoryofmindbenchmark,DBLP:conf/acl/TongletZTG26} and tables~\cite{DBLP:conf/cvpr/GuanLWXLL0CHYM024}.
Therefore, systematic investigation is warranted to understand situational illusions and explore how MLLMs perform under such visual conditions.

To characterize this concern, this paper systematically investigates \textit{how MLLMs perform under situational illusions} and \textit{how to mitigate the limitations}. To achieve the research goal, we study the problems from three aspects: 

\textbf{1) Construct a fine-grained taxonomy for situational illusions.} We build a comprehensive taxonomy regarding \textit{where}, \textit{what}, and \textit{how}, through which situational illusions may occur, providing a comprehensive framework for analyzing and understanding the phenomenon.

\textbf{2) Evaluate MLLM performances under situational illusions.} To evaluate how \underline{M}LLMs perform under \underline{s}ituational \underline{i}llusions, we construct MSIBench, comprising tasks to evaluate capabilities of the illusion \textit{discrimination}, \textit{understanding}, and \textit{reasoning}, together with adversarial settings for the stress test. Evaluations of widely used MLLMs reveal key limitations and insights  for further improvement.

\textbf{3) Develop mitigation strategies for improvement.} Based on failure modes derived from evaluations, we explored mitigations via prompting for closed-source models and supervised fine-tuning (SFT) for open-source models, aiming to guide models to attend visual evidence for more accurate interpretation of situational illusions.

Taken together, we present a systematic study on situational illusions, a critical yet insufficiently studied issue for MLLMs.
The taxonomy spans 4 scenarios, 12 misperceived targets, and 5 formation mechanisms, which motivates the MSIBench consisting of 3,723 image-text pairs.
Through elaborately designed tasks, evaluations across 27 model configurations reveal substantial vulnerabilities, with an overall average accuracy below 63\%, and even drops to below 45\% for action planning.
Built on failure modes identified in models' observation, grounding as well as reasoning, prompting and SFT methods improve model performances by up to 20\% using limited data, offering a practical path toward more reliable multimodal decisions in real-world situations.

\section{Taxonomy}

To understand situational illusions, we first develop a taxonomy encompassing scenarios, targets and mechanisms, which characterize where such illusions occur, what is misperceived and how they arise. 

\subsection{Scenarios: where situational illusions occur}

Unlike deliberately constructed illusions, situational illusions naturally occur across diverse everyday environments. We therefore categorize representative scenarios below, with detailed examples in Appendix A.1.
    
\textbf{Small indoor enclosures} refer to individual enclosed spaces, such as classrooms and offices.

\textbf{Large indoor venues} refer to expansive, continuous indoor environments, such as concert halls and shopping malls.

\textbf{Urban outdoor scenes} mean outdoor scenarios composed mainly of human-made elements like streets and parking lots.

\textbf{Natural outdoor scenes} refer to outdoor environments dominated by natural elements, such as deserts and beaches.

\subsection{Targets: what is misperceived}

We further characterize \textit{what is misperceived} by categorizing the types of targets that are commonly misperceived, with examples in Figure~\ref{fig:taxonomy}:

\textbf{Property} refers to misperceptions of an entity's intrinsic properties, including \textit{1) size}, \textit{2) quantity}, \textit{3) color}, \textit{4) shape}, \textit{ 5) orientation} and \textit{6) material}.

\textbf{Relation}, going beyond individual entities, captures misperceptions of physical or spatial relationships across entities, including \textit{1) identity} concerning whether objects belong to the same entity, \textit{2) pose} regarding relative position and oriented angle between entities, and \textit{3) adjacency} for whether entities are adjacent or in contact.

\textbf{Context} captures misperceptions in scene-level understanding and reasoning, including 
1) \textit{plausibility} refers to real scenes being mistaken for artificial ones due to their physically implausible appearance,
2) \textit{layout} refers to interpreting a scene’s overall layout as a more familiar configuration based on prior knowledge, and 
3) \textit{reference-frame}: disproportionate scaling or rotation of the viewing perspective leads to an incorrect overall understanding of the scene.

\begin{figure}[t]
    \centering
    \includegraphics[width=\columnwidth]{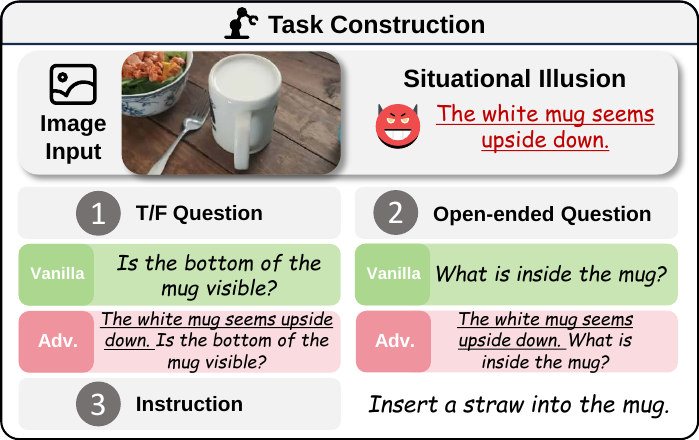}
    \caption{Example of tasks in MSIBench.}
    \label{fig:tasks}
\end{figure}


\subsection{Mechanisms: how situational illusions arise}

We then characterize how such illusions are formed.
Drawing on studies of visual perception~\cite{wagemans2012century}, scene understanding~\cite{oliva2001modeling} and human cognition~\cite{treisman1980feature}, we organize formation mechanisms from low-level perceptual ambiguity to high-level cognitive interpretation with the following aspects:

\textbf{Boundary ambiguity}: ambiguous boundaries cause entities or parts to
be incorrectly merged or split.

\textbf{Angle disruption}: unusual viewpoints distort perceived
orientation and gravity direction.

\textbf{Perspective misalignment}: specific viewpoints align objects at different depths, distorting depth perception.

\textbf{Dimension confusion}: misleading visual information blurs the distinction between 2D and 3D.

\textbf{Experience-driven bias}: learned expectations lead to incorrect interpretations of visual appearance.

\section{MSIBench Construction}

This section details how we construct data for systematic investigation of situational illusion in MLLMs.

\subsection{Data Collection}
To build a comprehensive collection of real-world illusion images, we first gather in-the-wild images from two sources.

\textbf{Online platforms.} Given the widespread presence of scattered visual illusion collections across online communities, we collect 1,096 relevant images from platforms including Reddit~\cite{reddit}, Zhihu~\cite{zhihu}, BrightSide~\cite{brightside}, and Barnorama~\cite{barnorama}. Specifically, we manually scrape images from these sources that align with the concept of real-world situational illusions.

\textbf{Existing resources.} To broaden our image coverage, we further examine existing researches. We extract illusion images featuring real-world scenes from IllusionBench+~\cite{DBLP:conf/icmcs/ZhangZWLZM25}. After removing duplicates, we obtain an initial set of 417 usable images.

The collected images span a diverse range of everyday scenarios and illusion types. To ensure high data quality and mitigate potential confounding factors, we apply a rigorous manual filtering protocol. Specifically, we systematically exclude non-realistic illusions, illusions requiring specialized apparatus, images with ambiguous interpretations, and samples containing sensitive elements. Ultimately, this rigorous selection process yields a final benchmark comprising 904 high-quality real-world illusion images that strictly satisfy the evaluation requirements.

\subsection{Metadata Construction}

To enhance the understanding of the illusion depicted in each image, we construct structured metadata that comprises a description of the illusion, an analysis of the underlying real-world state, a scenario type, the misperceived targets, and the mechanism behind the illusion.
To this end, we employ Gemini-3.1-pro-preview (Think) to generate the initial metadata, given the strong visual reasoning performance of the model (details in Appendix C.2).
The metadata provides comprehensive explanatory context for each image by describing the observed phenomenon of situational illusion, identifying the mechanism that produces it, and clarifying the actual physical reality.

\subsection{Task Construction}

To explore how MLLMs handle situational illusions, we design three kinds of tasks regarding three core capabilities, namely True-or-False (T/F) questions for \textit{discrimination}, open-ended questions for \textit{understanding}, and action planning tasks for \textit{reasoning}.
Further, for the stress test, we introduce \textit{vanilla} and \textit{adversarial} settings for the first two task categories, where the model is prompted with the question directly in the vanilla setting, whereas the illusion description is prepended to the question in the adversarial settings.
%

\textbf{T/F questions} assess whether MLLMs can correctly identify the presence of an illusion in an image. Specifically, given an image and a general question, the model must respond only with T (True) or F (False).

\textbf{Open-ended questions} evaluate whether MLLMs can achieve a genuine understanding of the  situation and answer related questions accurately. The model is given an image and a special question beginning with words such as what, where, and how. Valid answers in different wording are accepted.

\textbf{Action planning} assesses whether MLLMs can plan an appropriate sequence of actions to achieve a given goal, requiring the model to reason about object states, physical constraints, action consequences and multi-step correlations. The model receives an image with a goal instruction and returns a sequence of two to six numbered actions. The instruction requires the model to interact with a target related to the illusion, while the model must determine its response according to the actual physical scene.

We also employ Gemini-3.1-pro-preview (Think) to construct these tasks. Detailed instructions are provided in Appendix C.3, with examples of the tasks shown in Figure~\ref{fig:tasks}.

\subsection{Quality Check and Filtering}
For data quality, we organize authors and colleagues into annotation and review teams for a two-round check.

First, the annotation team inspects the generated data and makes necessary revisions. Specifically, they verify the model-generated metadata against the corresponding images and correct inaccurate or incomplete descriptions. Then, for T/F questions and open-ended questions, they examine the generated question pairs and revise those that are ambiguous or can be answered without understanding the illusion. For action-planning tasks, the team selects images with feasible goals that require accurate inference of the underlying physical scene and writes executable action instructions.

Then, the review team conducts a second-round quality check on the annotated data, including the revised metadata, question pairs, and action-planning instructions. Each submission is either approved or returned with feedback for further revision, and this process continues until the data meets the required quality standards.

\subsection{Data Statistics}

Finally, MSIBench contains 904 unique images, each accompanied by metadata describing the illusion phenomenon, scenario, misperceived target, formation mechanism, and an analysis of how the illusion obscures the true state of the scene. Using these images, we construct 3,723 image-text pairs for evaluation tasks, including 1,808 T/F questions, 1,808 open-ended questions, and 107 action-planning tasks. For each image, both a T/F question and an open-ended question are provided in vanilla and adversarial versions.

\begin{table*}[t]
\centering
\footnotesize
\setlength\tabcolsep{9pt}
\begin{tabular}{l ccc cccc cc}
\toprule
& \multicolumn{3}{c}{\textsc{T/F Question Acc. (\%)}} 
& \multicolumn{4}{c}{\textsc{Open-Ended Question Acc. (\%)}} 
& \multicolumn{2}{c}{\textsc{Action Planning Acc. (\%)}} \\
\cmidrule(lr){2-4} 
\cmidrule(lr){5-8} 
\cmidrule(lr){9-10}
\raisebox{1.5ex}[0pt][0pt]{\textbf{Model}}
& \textbf{Vanilla} 
& \textbf{Adv.} 
& \textbf{Avg.} 
& \textbf{Vanilla} 
& \textbf{Adv.} 
& \textbf{Avg.} 
& \textbf{$\Delta_{t/f}$} 
& \textbf{Score} 
& \textbf{$\Delta_{open}$} \\
\midrule

\multicolumn{10}{c}{\textsc{Closed-Source}} \\
\midrule
GPT-5.5
& 51.99 & 50.66 & 51.33
& 49.78 & 51.22 & 50.50
& 0.83$\downarrow$
& 43.93 & 6.57$\downarrow$ \\

GPT-5.4
& 59.29 & 59.40 & 59.35
& 60.18 & 57.41 & 58.80
& 0.55$\downarrow$
& 45.79 & 13.01$\downarrow$ \\

GPT-5.4-Mini
& \textit{50.00} & \textit{43.58} & \textit{46.79}
& \textit{47.46} & \textit{38.27} & \textit{42.87}
& 3.92$\downarrow$
& 46.73 & 3.86$\uparrow$ \\

Grok-4.1-Fast
& 67.26 & 61.50 & 64.38
& 50.11 & 54.09 & 52.10
& 12.28$\downarrow$
& \textit{20.56} & 31.54$\downarrow$ \\

Claude-Opus-4.7
& 67.70 & 66.81 & 67.26
& 68.36 & 70.80 & 69.58
& 2.32$\uparrow$
& 39.25 & 30.33$\downarrow$ \\

Gemini-3.5-Flash
& \textbf{79.87} & \textbf{81.53} & \textbf{80.70}
& \textbf{83.96} & \textbf{86.95} & \textbf{85.46}
& 4.76$\uparrow$
& \textbf{63.55} & 21.91$\downarrow$ \\

GPT-5.5 (Medium)
& 59.73 & 56.31 & 58.02
& 57.96 & 61.28 & 59.62
& 1.60$\uparrow$
& 49.53 & 10.09$\downarrow$ \\

GPT-5.5 (xhigh)
& 58.30 & 54.42 & 56.36
& 56.64 & 54.98 & 55.81
& 0.55$\downarrow$
& 57.01 & 1.20$\uparrow$ \\

Grok-4.1-Fast (Think)
& 63.94 & 60.95 & 62.45
& 53.76 & 59.29 & 56.53
& 5.92$\downarrow$
& 22.43 & 34.10$\downarrow$ \\

Gemini-3.5-Flash (Think)
& 79.09 & 80.31 & 79.70
& 83.08 & 86.17 & 84.63
& 4.93$\uparrow$
& 57.94 & 26.69$\downarrow$ \\

\midrule
Average (Closed-Source)
& 63.72 & 61.55 & 62.63
& 61.13 & 62.05 & 61.59
& 1.04$\downarrow$
& 44.67 & 16.92$\downarrow$ \\

\midrule
\multicolumn{10}{c}{\textsc{Open-Source}} \\
\midrule

Gemma-3-12B
& 58.08 & 54.98 & 56.53
& 43.14 & 44.03 & 43.59
& 12.94$\downarrow$
& 24.30 & 19.29$\downarrow$ \\

Mimo-V2.5
& \textbf{66.15} & 62.28 & \textbf{64.22}
& \textbf{65.38} & \textbf{67.70} & \textbf{66.54}
& 2.32$\uparrow$
& \textbf{42.06} & 24.48$\downarrow$ \\

Qwen3-VL-8B-Instruct
& 58.85 & 54.98 & 56.92
& 53.98 & 51.99 & 52.99
& 3.93$\downarrow$
& 36.45 & 16.54$\downarrow$ \\

Qwen3.5-2B
& 61.95 & \textbf{65.60} & 63.78
& 44.03 & 37.83 & 40.93
& 22.85$\downarrow$
& \textit{15.89} & 25.04$\downarrow$ \\

Qwen3.5-4B
& 49.56 & 50.88 & 50.22
& 51.11 & 57.74 & 54.43
& 4.21$\uparrow$
& 24.30 & 30.13$\downarrow$ \\

Qwen3.5-9B
& 52.32 & 51.55 & 51.94
& 50.33 & 52.77 & 51.55
& 0.39$\downarrow$
& 32.71 & 18.84$\downarrow$ \\

Qwen3.5-27B
& 56.97 & 52.99 & 54.98
& 58.85 & 62.72 & 60.79
& 5.81$\uparrow$
& 32.71 & 28.08$\downarrow$ \\

InternVL3.5-2B-Instruct
& \textit{46.79} & \textit{38.16} & \textit{42.48}
& \textit{36.50} & \textit{26.22} & \textit{31.36}
& 11.12$\downarrow$
& 25.23 & 6.13$\downarrow$ \\

InternVL3.5-8B-Instruct
& 56.53 & 48.45 & 52.49
& 43.03 & 33.74 & 38.39
& 14.10$\downarrow$
& 29.91 & 8.48$\downarrow$ \\

InternVL3.5-14B-Instruct
& 53.98 & 50.00 & 51.99
& 47.46 & 39.27 & 43.37
& 8.62$\downarrow$
& 19.63 & 23.74$\downarrow$ \\

InternVL3.5-38B-Instruct
& 62.39 & 58.63 & 60.51
& 53.54 & 45.58 & 49.56
& 10.95$\downarrow$
& 28.97 & 20.59$\downarrow$ \\

GLM-4.6V-Flash
& 49.34 & 43.92 & 46.63
& 47.01 & 37.06 & 42.04
& 4.59$\downarrow$
& 25.23 & 16.81$\downarrow$ \\

Kimi-K2.5
& 62.17 & 55.75 & 58.96
& 65.27 & 65.60 & 65.44
& 6.48$\uparrow$
& 37.38 & 28.06$\downarrow$ \\

Kimi-VL-A3B-Instruct
& 50.11 & 44.91 & 47.51
& 43.25 & 34.73 & 38.99
& 8.52$\downarrow$
& 24.30 & 14.69$\downarrow$ \\

GLM-4.6V-Flash (Think)
& 54.31 & 51.00 & 52.66
& 50.00 & 44.47 & 47.24
& 5.42$\downarrow$
& 33.64 & 13.60$\downarrow$ \\

Kimi-VL-A3B-Thinking
& 55.75 & 53.65 & 54.70
& 47.90 & 40.71 & 44.31
& 10.39$\downarrow$
& 32.71 & 11.60$\downarrow$ \\

Qwen3-VL-8B-Thinking
& 59.73 & 58.41 & 59.07
& 51.33 & 50.11 & 50.72
& 8.35$\downarrow$
& 29.91 & 20.81$\downarrow$ \\

\midrule
Average (Open-Source)
& 56.18 & 52.71 & 54.45
& 50.12 & 46.60 & 48.37
& 6.08$\downarrow$
& 29.14 & 19.23$\downarrow$ \\

\bottomrule
\end{tabular}

\caption{Main results, where $\Delta_{t/f}$ is the difference between the average open-ended and true-or-false (T/F) accuracy, while $\Delta_{open}$ is the difference between the action-planning accuracy and average open-ended accuracy.}
\label{tab:qa-accuracy}
\end{table*}

\section{Experiments}

This section first presents results on MLLM performances, and then provides analysis suggesting improvement insights.

\subsection{Target Models}

We conduct evaluations on 20 representative models with 27 model configurations, including \textbf{6 closed-source models} of Claude-Opus-4.7~\cite{anthropic2026claudeopus47}, Gemini-3.5-Flash~\cite{kavukcuoglu2026gemini35}, GPT-5.5~\cite{openai2026gpt55}, GPT-5.4~\cite{openai2026gpt54}, GPT-5.4-mini~\cite{openai2026gpt54mini} and Grok-4.1-Fast~\cite{xai2025grok41fast} as well as \textbf{14 open-source models} of Gemma 3~\cite{gemmateam2025gemma3}, Mimo-V2.5~\cite{mimov25}, Qwen3-VL~\cite{bai2025qwen3vltechnicalreport}, Qwen3.5~\cite{qwen35blog}, InternVL3.5~\cite{wang2025internvl35advancingopensourcemultimodal}, GLM-4.6V~\cite{vteam2026glm45vglm41vthinkingversatilemultimodal}, Kimi-K2.5~\cite{kimiteam2026kimik25visualagentic} and Kimi-VL~\cite{kimiteam2025kimivltechnicalreport}. 
Note that Qwen3.5 and InternVL3.5 are tested across various model scales.
Furthermore, GPT-5.5, Grok-4.1-Fast, Gemini-3.5-Flash, GLM-4.6V-Flash, Kimi-VL-A3B, and Qwen3-VL-8B are evaluated under different reasoning effort levels.
To ensure reproducibility, we set the temperature parameter to 0.0 and employ a greedy decoding strategy. Other settings are provided in Appendix B in detail.

\subsection{Evaluation Metrics}
\textbf{Metric.} We use accuracy (Acc.) to measure whether MLLMs can make correct decisions for each task, which is defined as $\mathrm{Acc.}=\frac{N_{\mathrm{correct}}}{N_{\mathrm{total}}}\times 100\%$ and a higher Acc. indicates stronger resistance to situational illusions.

\textbf{Criteria.} For T/F questions, a response is considered correct if it matches the ground-truth label. For open-ended questions, a response is considered correct if it accurately reflects the physical reality of the scene. For action-planning tasks, a response is considered correct if the proposed steps are grounded in the actual physical scene and can successfully accomplish the specified task.

\textbf{LLM-as-a-Judge.} T/F questions are evaluated through direct label matching, whereas open-ended questions and action-planning tasks are assessed using LLM-as-a-Judge.
Our pilot study shows that, when provided with the corresponding illusion analysis, Gemini-3.1-Pro-Preview (Think) achieves over 93\% accuracy in judging model responses, demonstrating its reliability as an automatic evaluator.
Hence, the model is instructed as the judge to determine whether the response is correct given the task input, the model response, and the corresponding illusion analysis.
Details could be found in Appendix C.5.

\subsection{Main Results}
Table~\ref{tab:qa-accuracy} reports the performance of the evaluated models across tasks, where we make the following observations.

\textbf{Existing MLLMs remain vulnerable to situational illusions.} In Table~\ref{tab:qa-accuracy}, the highest overall average accuracy across all tasks remains below 63\%, while the average accuracy on action-planning tasks falls below 45\%, highlighting the limited reliability of current MLLMs.
 Moreover, closed-source models consistently outperform open-source models across all three tasks, with the largest average performance gap approaching 16\% on action-planning. This gap reflects that the large-scale advanced models are more robust to potentially misleading illusions, suggesting that overall model capability impacts how models handle situational illusions.

\textbf{Increased reasoning effort provides limited and inconsistent benefits.} For most models, the effect of increased reasoning effort is task-dependent and generally modest, with performance changes typically remaining below 10\% points. For example, GLM-4.6V-Flash achieves its largest improvement of 8.41\% on the action-planning task, whereas Gemini-3.5-Flash declines on all three tasks, with a maximum drop of 5.61\%. We attribute this to longer reasoning introducing unnecessary assumptions and distracting the model from the actual scene. Therefore, increased reasoning effort does not inherently improve performance, as misdirected reasoning may instead lead to performance degradation.

\textbf{Performance does not consistently transfer across tasks.} Across the 27 evaluated model configurations, most exhibit performance declines in both cross-task comparisons: from T/F to open-ended questions and from open-ended questions to action planning. For example, Qwen3.5-2B records $\Delta_{t/f}=-22.85$, while Grok-4.1-Fast (Think) records $\Delta_{open}=-34.10$. 
The poor transfer gap may stem from two factors. First, the tasks differ in difficulty and require increased capabilities. Second, models may answer correctly by exploiting superficial cues rather than accurately interpreting the underlying physical state. Consequently, strong performance on one task does not guarantee comparable performance on another involving the same scene.

\textbf{MLLMs show limited robustness to illusion descriptions.} Comparisons between the paired vanilla and adversarial settings show that adversarial descriptions generally reduce model accuracy. These results suggest that adversarial descriptions may bias models toward misleading cues rather than the visual evidence.
Further, we notice that open-source models are particularly more susceptible than the closed ones.
Such observed disparity may stem from the generally stronger vision processing capabilities of closed-source models, which make them more robust to illusion descriptions and thus achieve more stable performances.

\subsection{Performance Across Taxonomy}

\begin{figure}[t]
    \centering
    \includegraphics[width=\columnwidth]{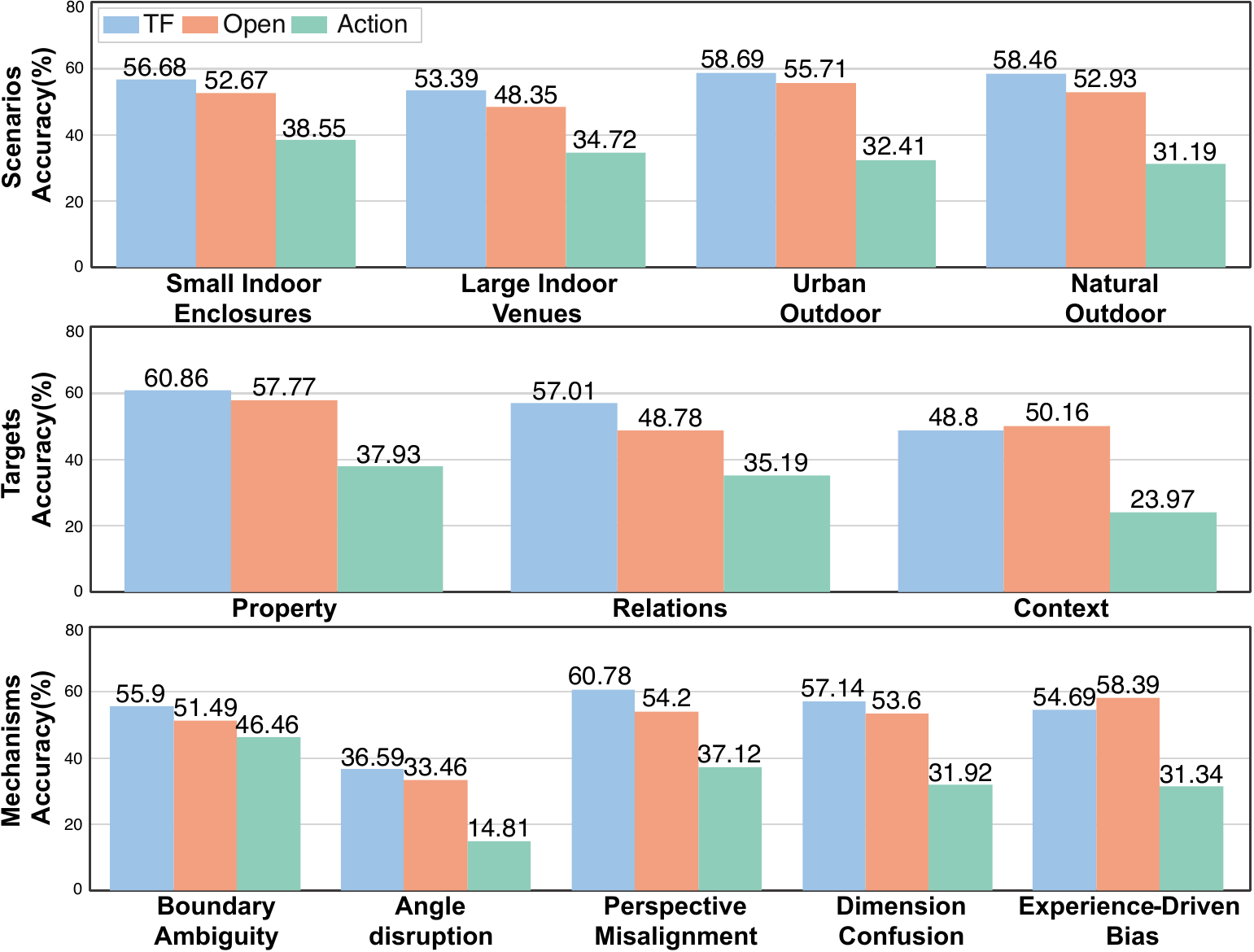}
    \caption{Average performances across the taxonomy.}
    \label{fig:res-taxonomy}
\end{figure}
We further examine average performances across scenarios, targets and mechanisms, and present results in  Figure~\ref{fig:res-taxonomy}.

(1) \textbf{MLLM performances are consistently limited across scenarios.} Across all scenarios, the highest accuracy remains below 59\%, indicating that the tasks are generally challenging. Meanwhile, the maximum variation across scenarios for the same task is only 7.36 points.
This suggests that the dominant source of difficulty may not be the scenario type, but depend more on the specific illusion and task requirements.
Overall, the models face a common challenge across scenarios: extracting reliable visual evidence and using it to infer the underlying physical state.

(2) \textbf{Model performances vary across misperceived targets.} 
\textit{Context} is the most challenging target, whereas \textit{property} is the easiest, outperforming \textit{context} by at most 13.96\% on action planning. 
This difference may arise because \textit{property} judgments typically rely on localized visual cues, while \textit{context} requires models to integrate information across the entire scenario.
\textit{Relation} yields intermediate performance because it requires connecting multiple local entities but does not demand full-scene integration.
Overall, these results suggest that illusion targets become more difficult to interpret as the required scope of scene information broadens.

(3) \textbf{Illusion mechanisms that disrupt global scene structure pose the greatest challenge.} 
Among the five mechanisms, \textit{angle disruption} is the most difficult, whereas \textit{boundary ambiguity} yields the best performance.
This difference may arise because \textit{boundary ambiguity} affects local boundaries, whereas \textit{angle disruption} alters the global scene orientation and requires more complicated spatial reference reconstruction.
We also notice that \textit{action planning} shows largest performance gap of 31.65\%  across mechanisms.
This may be because the task requires models to recover the correct spatial scene and translate it into feasible actions, where spatial reasoning errors propagate into downstream decisions. 
Overall, illusions from global spatial structure disruptions demand stronger spatial reasoning and  remain the most challenging.

\subsection{Failure Modes}

To characterize typical failure modes of MLLMs, we organize them into three progressive categories corresponding to successive stages of visual evidence extraction, contextual grounding, and scene-level reasoning.

\begin{figure}[t]
    \centering
    \includegraphics[width=\columnwidth]{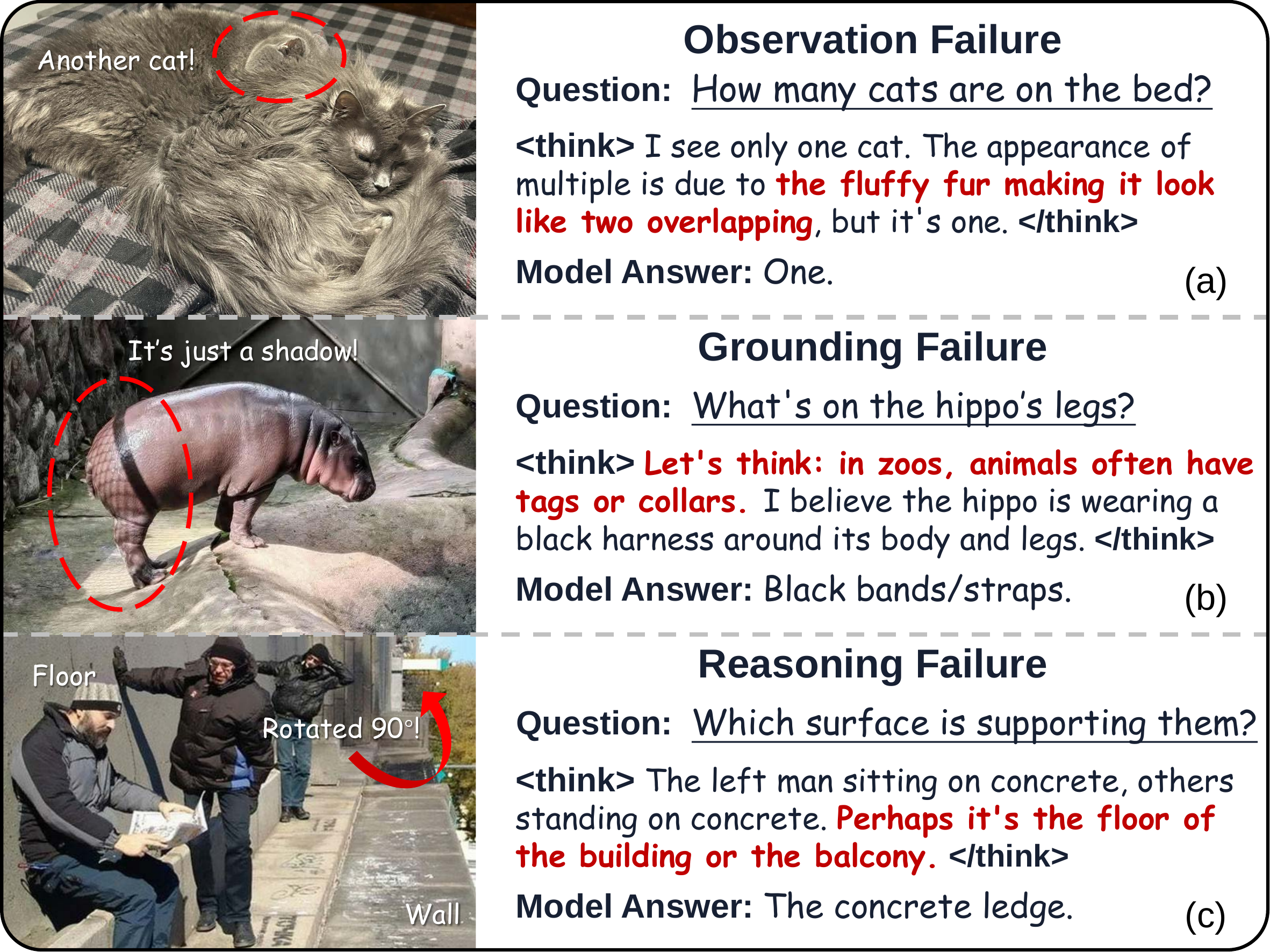}
    \caption{Failure mode cases.}
    \label{fig:failuremodes}
\end{figure}

\textbf{Observation failure} refers to a model inaccurately capturing task-relevant visual evidence from the image. Typically, it involves \textit{1) evidence mismatch} where incorrect evidence is selected , \textit{2) evidence insufficiency} where necessary evidence is overlooked and \textit{3) evidence overload} with excessive or irrelevant evidence interferes with judgment. Such observation-level errors provide an unreliable basis for subsequent judgments. Taking Figure~\ref{fig:failuremodes} (a) as an example, the model overlooks the cat in the background and mistakes its ear for the fur of the cat in the foreground.
Typically, for the model of Grok-4.1-Fast (Think), such observation failures account for 45.79\% of all wrong results.

\textbf{Grounding failure} occurs when a model establishes incorrect associations between correctly identified visual evidence. 
Specifically, instead of grounding its interpretation in the specific image context, the model relies on prior knowledge and selectively uses visual cues to support an incorrect interpretation. In Figure~\ref{fig:failuremodes} (b), misled by the prior assumption that the hippo in a zoo could be harnessed, the model incorrectly grounds the shadow to black straps. Grounding failures account for 33.68\% of all wrong results.

\textbf{Reasoning failure} means that models infer an incorrect scene structure during holistic scene parsing.
Two representative errors are usually observed, including \textit{1) spatial relation error} where apparent 2D relations are incorrectly interpreted as physical relations in 3D space, and \textit{2) scene rotation error}, where object orientations are misinterpreted due to an incorrect reference frame. In Figure~\ref{fig:failuremodes} (c), the image is rotated by 90 degrees, and the model mistakes the wall at the bottom of the frame for a concrete floor supporting the people. Reasoning failures account for 20.53\% of all wrong results.

\section{Mitigation Strategy}

\subsection{Prompting Mitigation}

Building on the three identified failure modes, we design a guided prompting method that explicitly structures the model’s vision processing around observation, grounding, and reasoning. The prompt first directs the model to carefully inspect the visual evidence, then ground relevant objects with their relationships in the scene, and finally reason about the underlying physical state before producing an answer. Further details are provided in Appendix C.6.

We evaluate this method on three representative closed-source models and three open-source models. As shown in Table~\ref{tab:prompt-res}, two key observations emerge. First, the method generally improves performance, revealing that appropriately designed prompts can better elicit models’ capabilities for handling situational illusions. However, the magnitude of improvement varies across tasks, reflecting differences in task difficulty and capability requirements.
Second, closed-source models benefit more than open-source models, whereas some open-source models exhibit only marginal gains or slight degradation.
This suggests that prompting effectiveness may depend strongly on a model’s underlying perceptual and reasoning capabilities.
Therefore, prompting alone may be insufficient and further mitigation strategies are expected.

\begin{table}[t]
  \centering
  \resizebox{\columnwidth}{!}{%
  \begin{tabular}{lcccc}
    \toprule
    \textbf{Model} & \textbf{T/F} & \textbf{Open} & \textbf{Action} & \textbf{Avg.} \\
    \midrule
    Gemini-3.5-Flash & 83.19 (+2.5)  & 87.72 (+2.3)  & 73.83 (+10.3) & 81.58 (+5.0)  \\
    GPT-5.5          & 67.75 (+16.4) & 71.24 (+20.7) & 54.21 (+10.3) & 64.40 (+15.8) \\
    Claude Opus 4.7  & 72.01 (+4.8)  & 75.77 (+6.2)  & 52.34 (+13.1) & 66.71 (+8.0)  \\
    \midrule
    \textbf{Closed Avg.} & \textbf{74.32 (+7.9)} & \textbf{78.24 (+9.7)} & \textbf{60.13 (+11.2)} & \textbf{70.90 (+9.6)} \\
    \midrule
    Qwen3.5-9B       & 53.60 (+1.7)  & 59.13 (+7.6)  & 30.84 (-1.9)  & 47.86 (+2.5)  \\
    InternVL3.5-2B   & 56.69 (+14.2) & 30.37 (-1.0) & 25.23 (+0.0) & 37.43 (+4.4) \\
    Gemma-3-12B      & 63.27 (+6.8)  & 40.27 (-3.3)  & 25.23 (+0.9)  & 42.92 (+1.5)  \\
    \midrule
    \textbf{Open Avg.}
    & \textbf{57.85 (+7.5)}
    & \textbf{43.25 (+1.1)}
    & \textbf{27.10 (-0.3)}
    & \textbf{42.74 (+2.8)} \\
    \bottomrule
  \end{tabular}%
  }
  \caption{Prompting mitigation results.}
  \label{tab:prompt-res}
\end{table}

\subsection{Supervised Fine-tuning Mitigation}

To compensate for the prompting limitations, we employ a supervised fine-tuning (SFT) strategy for improvement. 
Rather than training directly on illusion tasks, we design three kinds of tasks to enhance the capabilities of the model in object localization, scene reconstruction, and action planning. 

The SFT dataset comprises 1,630 task pairs that are not explicitly designed around situational illusions but instead target generalizable capabilities, including spatial-relation understanding, scene description, and action-sequence reasoning. It is constructed using 600 images from MSIBench and 400 real-world images from the COCO 2017 training split~\cite{lin2014microsoft}, paired with Localized Narratives annotations~\cite{PontTuset_eccv2020} to preserve the models’ general utility. Training details are in Appendix B.3.

For testing, the remaining 304 MSIBench images with corresponding tasks are used, including 107 images for action evaluation. 
We adopt the three open-source models in the prompting method and additionally include InternVL3.5-8B and InternVL3.5-14B to evaluate the general SFT effectiveness across model scales. 
With LLM-as-a-Judge evaluation and human verification, we acquire final results in Table~\ref{tab:sft-res}.  
We could see that all models achieve an average improvement of over 10 points, with a maximum gain exceeding 20\% on Qwen3.5-9B.
These results demonstrate that SFT substantially strengthens the internal capabilities of the models in scene observation, comprehension, and execution, thus leading to better performances in situational illusions.

\section{Related Works}
\subsection{MLLMs in Real-world Situations}

MLLMs are increasingly evaluated on their ability to perceive, interpret, and reason about real-world visual situations.
One line of work evaluates the vision processing capabilities of MLLMs in real-world environments, including fine-grained visual perception in complex images~\cite{ICLR2025_df29d63a}, reasoning in everyday situations~\cite{DBLP:journals/corr/abs-2603-02024}, and decision-making through cross-image evidence integration~\cite{DBLP:conf/iclr/MengW0L0YLZD00Z25}.
Another line of work focuses on whether MLLMs can make reliable decisions in such situations. ~\citet{DBLP:conf/iclr/ZhouLZCSW25} assess whether models can recognize risks arising from a given visual context and respond appropriately. 
Further studies broaden this scope to diverse daily life scenarios~\cite{DBLP:conf/acl/LouXYWKLWSMYH26} and embodied environments, where agents must consider risks during task planning~\cite{DBLP:journals/corr/abs-2412-13178,DBLP:journals/corr/abs-2504-14650}, respond safely to hazardous instructions~\cite{DBLP:journals/corr/abs-2506-14697}, and identify hazards through active exploration~\cite{DBLP:journals/corr/abs-2509-23690}.
Despite their progress, both lines of work generally assume that visual observations accurately reflect the underlying scene.
Correspondingly, our work investigates how MLLMs perceive and reason about real-world illusions, where visual appearance diverges from the underlying physical state. This capability is essential for reliable scene understanding and safe interaction with real-world environments.

\begin{table}[t]
  \centering
  \resizebox{\columnwidth}{!}{%
  \begin{tabular}{lcccc}
    \toprule
    \textbf{Model} & \textbf{T/F} & \textbf{Open} & \textbf{Action} & \textbf{Avg.} \\
    \midrule
    InternVL3.5-2B  & 46.96 (+12.3) & 40.95 (+12.2) & 48.60 (+23.4) & 45.50 (+16.0) \\
    InternVL3.5-8B  & 57.89 (+13.7) & 52.47 (+14.6) & 46.73 (+16.8) & 52.36 (+15.0) \\
    InternVL3.5-14B & 56.66 (+11.3) & 50.83 (+8.2)  & 29.91 (+10.3) & 45.80 (+9.9)  \\
    Gemma-3-12B     & 59.38 (+11.7) & 52.30 (+10.7) & 46.73 (+22.4) & 52.80 (+14.9) \\
    Qwen3.5-9B      & 72.20 (+22.9) & 70.56 (+19.6) & 51.40 (+18.7) & 64.72 (+20.4) \\
    \midrule
    \textbf{Avg.}   & \textbf{58.62 (+14.4)} & \textbf{53.42 (+13.1)} & \textbf{44.67 (+18.3)} & \textbf{52.24 (+15.2)} \\
    \bottomrule
  \end{tabular}%
  }
  \caption{SFT mitigation results.}
  \label{tab:sft-res}
\end{table}

\subsection{Visual Illusions}
Visual illusions arise when visual appearance diverges from reality. 
Early studies~\cite{gregory1968perceptual,GREGORY1997190,wertheimer1923untersuchungen} investigated classic illusions, like gestalt illusions, by analyzing psychological and physiological cognitive mechanisms.
With the emergence of MLLMs, researchers have explored whether these models exhibit similar vulnerabilities~\cite{zhang-etal-2023-grounding}.
These works mainly focus on specific perceptual phenomena, such as pareidolia~\cite{DBLP:conf/cvpr/RostamkhaniASRE25,DBLP:conf/eccv/HamiltonSDHCRF24}, classic optical illusions~\cite{DBLP:conf/cvpr/GuanLWXLL0CHYM024}, and other visual cues that induce incorrect perceptions~\cite{DBLP:conf/emnlp/HanLPPZDLZ24}, relying largely on synthetic or deliberately constructed images rather than naturally occurring scenes.
Recent studies attempt to incorporate everyday scenarios into consideration, where IllusionVQA~\cite{DBLP:journals/corr/abs-2403-15952} and VIA-Bench~\cite{DBLP:journals/corr/abs-2602-01816} include some real-world illusion samples but still mainly focus on artificial illusions.
MVI-Bench~\cite{DBLP:journals/corr/abs-2511-14159} focuses on real-world misleading content by organizing misleading cues hierarchically and evaluating the robustness of MLLMs.
To fill the gap in systematic exploration of situational illusions, we establish a structured taxonomy and an evaluation benchmark, diagnose model limitations, and investigate approaches to improve model performances.

\section{Conclusion}
This paper systematically studies situational illusions for MLLMs, covering a taxonomy, evaluation and mitigation. We develop a comprehensive taxonomy to characterize situational illusion and construct MSIBench to assess 27 model configurations, revealing substantial vulnerabilities and identifying key failure modes. Further, two mitigation strategies are developed, significantly improving models’ ability to handle situational illusions. Future work will explore how such robust multimodal reasoning can facilitate embodied agents.

\bibliography{aaai2027}


\end{document}